\documentclass[10pt,twocolumn,letterpaper]{article}

\usepackage{iccv}              

\definecolor{iccvblue}{rgb}{0.21,0.49,0.74}
\usepackage[pagebackref,breaklinks,colorlinks,allcolors=iccvblue]{hyperref}

\usepackage{mathtools}
\usepackage{multirow}
\usepackage{algorithm}
\usepackage{algpseudocode}
\usepackage{graphicx}
\usepackage{caption}
\usepackage{subcaption}
\newcommand{\bv}[1]{\boldsymbol{\mathbf{#1}}}

\def\paperID{11717} 
\def\confName{ICCV}
\def\confYear{2025}

\title{Attention to Neural Plagiarism: \\ Diffusion Models Can Plagiarize Your Copyrighted Images!}

\author{
Zihang Zou\textsuperscript{1} \quad Boqing Gong\textsuperscript{2} \quad Liqiang Wang\textsuperscript{1} \\
\textsuperscript{1}University of Central Florida\\
\textsuperscript{2}Boston University\\
{\tt\small \{zihang.zou, liqiang.wang\}@ucf.edu, bgong@bu.edu}
}

\begin{document}
\twocolumn[{
\renewcommand\twocolumn[1][]{#1}
\maketitle
\begin{center}
    \centering
    \captionsetup{type=figure}
    \includegraphics[width=0.9\linewidth]{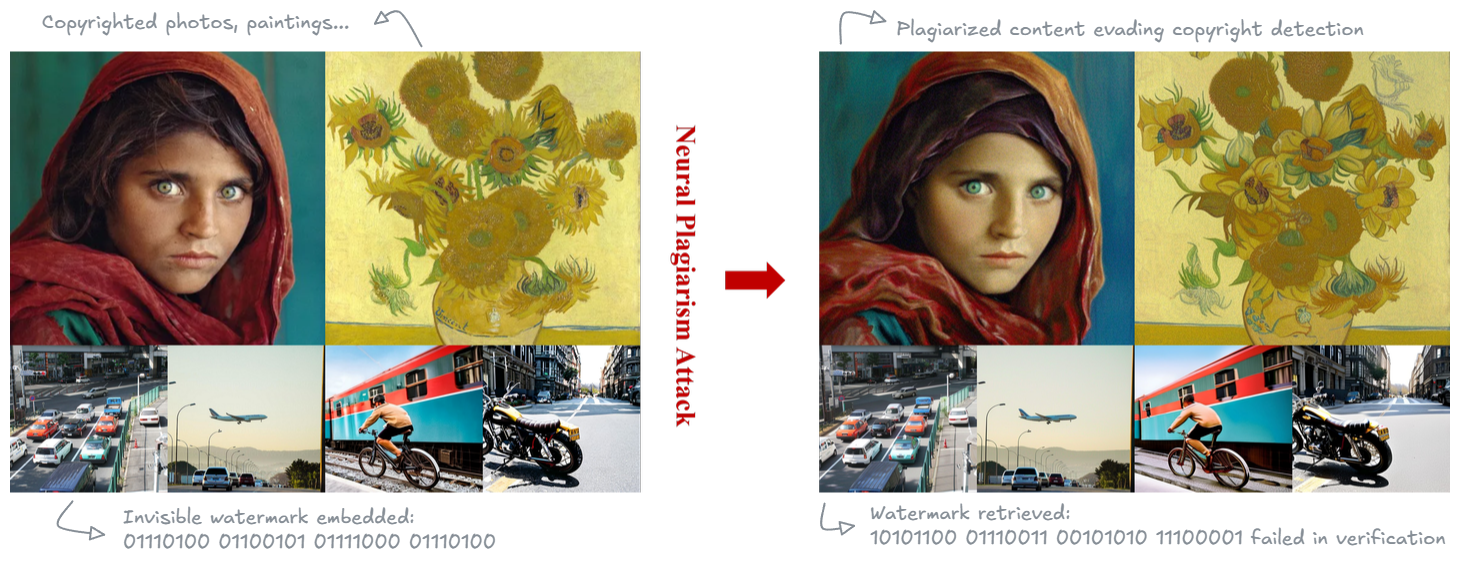}
    \captionof{figure}{Copyrighted images (top left) and watermarked data (bottom left) can be easily plagiarized by diffusion models.}
\end{center}
}]

\begin{abstract}
In this paper, we highlight a critical threat posed by emerging neural models—data plagiarism. We demonstrate how modern neural models (\eg, diffusion models) can effortlessly replicate copyrighted images, even when protected by advanced watermarking techniques. To expose the vulnerability in copyright protection and facilitate future research, we propose a general approach regarding neural plagiarism that can either forge replicas of copyrighted data or introduce copyright ambiguity. Our method, based on ``anchors and shims'', employs inverse latents as anchors and finds shim perturbations that can gradually deviate the anchor latents, thereby evading watermark or copyright detection. By applying perturbation to the cross-attention mechanism at different timesteps, our approach induces varying degrees of semantic modifications in copyrighted images, making it to bypass protections ranging from visible trademarks, signatures to invisible watermarks. Notably, our method is a purely gradient-based search that requires no additional training or fine-tuning. Empirical experiments on MS-COCO and real-world copyrighted images show that diffusion models can replicate copyrighted images, underscoring the urgent need for countermeasures against neural plagiarism. 
Source code is available at: \href{https://github.com/zzzucf/Neural-Plagiarism}{https://github.com/zzzucf/Neural-Plagiarism}.
\end{abstract}

\section{Introduction}
Plagiarism has long been a critical concern for data copyrights,
and recent advances in generative models\cite{rombach2022high, zhang2023adding, ramesh2021zeroshot, ramesh2022hierarchical, saharia2022imagen} have intensified this issue. These neural generative models, trained on vast datasets, are capable of producing high quality images, showing a troubling potential for what we term ``neural plagiarism''. Specifically, they may memorize training data and then directly reproduce copyrighted content~\cite{carlini2021extracting, carlini2023extracting}, or generate outputs that closely resemble copyrighted works.
As generative AI continues to evolve, the risk of copyright infringement escalates across a broad spectrum of data types, including watermarked user's data, photographs, artworks, signatures, and human portraits.

Legal regulations have been put in place to counteract the misconducts of neural models. For example, the General Data Protection Regulation (GDPR)~\cite{gdpr} prohibits neural models to train on user's data without consent; traditional copyright laws~\cite{usdmca} and trademark laws~\cite{usptoTrademarkLaw} restrict generative AI from producing copyrighted images. 

Beyond legal measures, technical approaches have also been implemented to protect intellectual property. Watermarking has proven to be an effective tool for copyright protection~\cite{hamidi2021dct, zhang2019robust}, particularly in digital media. Visual watermarks, such as trademarks and signatures, are widely used to assert ownership and deter unauthorized use of images. Additionally, emerging invisible watermarking techniques~\cite{hamidi2021dct, zhang2019robust, zhu2018hidden} enable content creators to embed robust copyright protections within digital assets without noticeably degrading image quality.


Despite these legal frameworks and technical efforts, diffusion models show a great potential to plagiarism as it can effortlessly generate replicas~\cite{zhao2023invisible, an2024benchmarking, liu2025image} of copyrighted data while circumventing protection measures like watermarking~\cite{zhao2023invisible, jiang2023evading, liu2024image, an2024benchmarking}.

In this work, we focus on plagiarism of copyrighted images from diffusion models. 
We propose a novel ``shim'' and ``anchor'' method for diffusion models to enable optimization in latent space where we can gradually modify semantic information to bypass copyright protections. We also present a novel perturbation technique based on attention, where we aim to find the alternative query, key and value that generate outputs as original ones to maintain the semantic consistency. 


In summary, our main contributions are as follow:
\begin{itemize}
    \item Introduce a general neural plagiarism pipeline for diffusion models that exposes vulnerabilities in existing data copyright protection methods and hence facilitate the development of more effective copyright protections.
    \item Develop a controllable search-based method and introduce a novel attention-based perturbation technique for diffusion models that enables a coarse-to-fine exploration of the semantic space with low memory consumption.
    \item Present a comprehensive analysis of forgery and ambiguity attacks that can lead to plagiarism in real-world cases, ranging from copyrighted artworks, signature, trademark to invisible watermarked data.
\end{itemize}

\section{Related Work}
\noindent\textbf{Intellectual Property and Traditional Watermarking.} Intellectual property (IP) is fundamental to human society—it incentives creativity, rewards innovation, and preserves culture. For example, Disney’s IP , such as Mickey Mouse, The Lion King, Frozen, is strictly protected under copyright laws~\cite{usptoTrademarkLaw} to prevent unauthorized use and plagiarism. Beyond legal protections, watermarking has long been an effective method for securing copyrighted content. Traditional techniques achieve this by embedding watermarks through geometric transformations in the frequency domain~\cite{hamidi2021dct, navas2008dwt}. These techniques create invisible watermark yet remain vulnerable  to image distortions. 

With recent development in artificial intelligence, deep learning models have emerged for robust watermark embedding. HiDDeN~\cite{zhu2018hidden} introduces an encoder–decoder framework while leveraging adversarial training~\cite{goodfellow2014generative} to improve watermark resilience. Subsequent studies explore self-supervised learning~\cite{zhang2019robust} to embed watermarks in latent spaces, employ attention mechanisms~\cite{fernandez2022watermarking}, and utilize differentiable simulations of real-world distortions~\cite{tancik2020stegastamp} to further enhance robustness of watermarking.

\noindent\textbf{Copyright Protection and Threats from Generative AI.} 
Recent advancements in diffusion models~\cite{sohl2015deep, rombach2022high, ho2020denoising, song2020denoising, peebles2023scalable} have ushered in a new era of generative AI, enabling the creation of highly realistic images, videos, and sounds that are virtually indistinguishable from reality. However, these breakthroughs have also raised concerns over misuse, such as deepfakes~\cite{thies2016face2face, suwajanakorn2017synthesizing, fried2019text}. In response, legal frameworks now require AI-generated content to be clearly labeled with a visible watermark.

To comply with legal requirements and protect copyrights, researchers are developing watermarking techniques specific to generative models, in particular, diffusion models. For instance, Stable Signature~\cite{fernandez2023stable} fine-tunes a VAE decoder to embed watermarks within images, which can then be extracted using a pretrained message decoder from HiDDeN~\cite{zhu2018hidden}. In comparison to traditional invisible watermark methods, Tree-Rings Watermark approach~\cite{tree_ring} injects a tree-ring–like watermark into the Fourier frequency domain of latent variables, generating semantically similar content while ensuring the latent watermark remains resilient to common image distortions. Building on this idea, Gaussian Shading~\cite{yang2024gaussian} employs a quantile-based method to mitigate the significant distribution shift in latent variables caused by the tree-ring watermark, achieving performance-lossless watermarking. Moreover, RingID~\cite{ci2025ringid} enhances the tree-ring approach with greater precision by incorporating multi-key identification through a series of augmented transformations.

Like a double-edged sword, these generative models unfortunately show greater potential for infringing copyrighted data. \textbf{Forgery attacks or regeneration attacks}, compromise data ownership by removing original watermarks~\cite{liu2023making, liu2024image, jiang2023evading, zhao2023invisible, saberi2023robustness}, making it difficult to trace content back to its source. 
In particular, Regen~\cite{zhao2023invisible} exploits diffusion models by adding noise in the forward process for a few step and then denoising via the reverse process, effectively removing most invisible watermarks~\cite{hamidi2021dct, zhang2019robust, fernandez2023stable}.
Rinse~\cite{an2024benchmarking} iterates this process multiple times to further improve watermark removal. Additionally, recent work~\cite{liu2024image} leverages ControlNet~\cite{zhang2023adding} and extra image encoder~\cite{oquab2023dinov2} to regenerate the watermarked images from clean noise under semantic constraints, thus removing latent watermarks like Tree-ring~\cite{tree_ring}.

Finally, \textbf{ambiguity attacks}~\cite{yuan2024ambiguity} remain an underestimated threat to copyright protection. In these attacks, replicas are generated with alternative watermarks linked to different ``original data'', creating competing claims of ownership and complicating the verification of the rightful owner. Despite their significant risks, existing defenses are limited, with the few available methods addressing only very specific scenarios~\cite{fan2019rethinking, zhang2020passport}. 
The only practical defense based on data creation timestamps is largely inapplicable, as creation times are often unrecorded in real-world cases.

\section{The Rising Threat ---``Neural Plagiarism''}\label{sec:problem_statement}
In this work, we focus on posing a new threat, denoted as neural plagiarism, which exploits neural models to \textit{replicate copyrighted data}, while confusing or evading existing copyright protection method. In this section, we formalize this problem into a scenario involved with a data owner, an attacker and a third-party verifier.

\noindent\textbf{Scope of Copyrighted Data.} We primarily focus on image data and address a broad range of data types subject to copyright protection in real-world scenarios. This includes traditional forms of copyrighted data, such as photographs, paintings, artworks and trademarks, as explicitly defined by copyright laws~\cite{usptoTrademarkLaw, usdmca}. Additionally, we consider user owned data protected by watermarking techniques.

\noindent\textbf{Data Owner's Capabilities.} Given an image data $x$, a data owner can choose the watermarking method along with watermark $w$ (\eg, a hidden bit stream~\cite{zhu2018hidden,tancik2020stegastamp}, a registered trademark or simply copyrighted images) and publishes only watermarked data $x^w$.

\noindent\textbf{Attacker's Ability.} An attacker only has access to the copyrighted data $x^w$ without knowing the watermark.

\noindent\textbf{The Plagiarism Attacks.} Formally, an attacker aims to develop a generative model $\mathcal{G}$ that replicates copyrighted data $x^w$, protected by a watermark $w$. The plagiarism occurs when the distance between the replicated $x^*$ and $x^w$ is small,
\begin{equation}
    x^* = \mathcal{G}(x^{w}) \text{ s.t. } d(x^*, x^{w}) < \delta
\end{equation}
Here, $d$ is a distance function (\eg, $l_2$ distance, perceptual distance~\cite{zhang2018unreasonable} or human justifications) and $\delta$ is a predefined threshold.

Given a third-party verifier $\mathcal{V}$, who can arbitrate the data ownership by recovering the watermark $w$ (\eg comparing bit accuracy, watermark pattern or copyrights violation), two major types of plagiarism attacks are considered:
\begin{itemize}
    \item\textbf{Forgery Attack:} An adversary leverages a generative model to replicate the copyrighted image while effectively removing the embedded watermark, which leads to a verification failure $\mathcal{V}(x^*) \neq w$.
    \item \textbf{Ambiguity Attack:} This attack removes the original watermark and creates ownership ambiguity by embedding an alternative watermark, causing problem in copyright arbitration: $\mathcal{V}(x^*) = w^*$ and $\mathcal{V}(x^w) = w$ while $x^*$ and $x^w$ are visually similar.
\end{itemize}

\section{Towards Neural Plagiarism Attacks}
Without loss of generality, this paper focuses on copyrighted images, including real-world copyrighted images and AI-generated images from diffusion models.



\subsection{Background}  
Diffusion models~\cite{sohl2015deep, rombach2022high, ho2020denoising, song2020denoising, peebles2023scalable}, originally inspired by thermodynamics, operate as denoising auto-encoders that iteratively remove noise from an image. In the \textbf{forward process}, noise is gradually injected into an initial image $\mathbf{x}_0$, eventually transforming it into an isotropic Gaussian. This process is defined as 
\[
q(\mathbf{x}_{1:T} \vert \mathbf{x}_0) = \prod_{t=1}^{T} q(\mathbf{x}_t \vert \mathbf{x}_{t-1}),
\]
where $q(\mathbf{x}_t \vert \mathbf{x}_{t-1}) = \mathcal{N}(\mathbf{x}_t; \sqrt{1 - \beta_t}\, \mathbf{x}_{t-1},\, \beta_t\mathbf{I})$ for each timestep. Using the reparameterization trick~\cite{kingma2013auto}, the noisy image at step $t$ can be expressed as
\begin{equation}
\mathbf{x}_t = \sqrt{\overline{\alpha}_t}\,\mathbf{x}_0 + \sqrt{1-\overline{\alpha}_t}\,\mathbf{\epsilon}    
\end{equation}

The \textbf{reverse (denoising) process} aims to recover the original image by progressively removing the injected noise. It is modeled as  
\[
p_\theta(\mathbf{x}_{0:T}) = p(\mathbf{x}_T) \prod_{t=1}^{T} p_\theta(\mathbf{x}_{t-1} \vert \mathbf{x}_t),
\]
with $p_\theta(\mathbf{x}_{t-1} \vert \mathbf{x}_t) = \mathcal{N}(\mathbf{x}_{t-1}; \boldsymbol{\mu}_\theta(\mathbf{x}_t, t),\, \boldsymbol{\Sigma}_\theta(\mathbf{x}_t, t))$.

In this formulation, the network learns to predict the mean \(\boldsymbol{\mu}_\theta(\mathbf{x}_t, t)\) and covariance \(\boldsymbol{\Sigma}_\theta(\mathbf{x}_t, t)\) that effectively reverse the forward diffusion. Recent advances, such as deterministic solvers~\cite{lu2022dpmSolver}, enable the \textbf{inversion process}~\cite{lu2022dpmSolver, hong2024exact, wallace2023edict} by iteratively reversing the deterministic reverse process, targeting the original latent variable that generates the image \(x\).

In practice, the reverse process is facilitated by a neural network based on the UNet architecture~\cite{ronneberger2015u}, denoted as $\epsilon_{\theta}$. This network is tasked with predicting the noise $\epsilon$ added at each diffusion step. Given a noisy image $\mathbf{x}_t$ and the corresponding timestep $t$, $\epsilon_{\theta}(\mathbf{x}_t, t)$ estimates the noise component, which is used to guide the denoising process. The network is trained by minimizing the mean squared error between the true noise and the predicted noise:
\[
\mathcal{L}_\text{simple} = \mathbb{E}_{t, \mathbf{x}_0, \epsilon} \Big[ \|\epsilon - \epsilon_{\theta}(\sqrt{\overline{\alpha}_t}\,\mathbf{x}_0 + \sqrt{1-\overline{\alpha}_t}\,\epsilon,\, t)\|^2 \Big].
\]

As a result, diffusion models are well-suited for data plagiarism. They can regenerate images through the inverse and denoising process~\cite{zhao2023invisible} and enable further optimization of semantic content via latent variables.

\subsection{Intuitive Objective Function}
The plagiarism attack aims to replicates a target copyrighted image \(\mathbf{x}^w_0\) by generating a visually similar attack output \(\mathbf{x}^*_0\). In diffusion models, latent variables at different timesteps capture various feature and may contain embedded watermarks (e.g., Tree-Ring~\cite{tree_ring} injects watermark on latent variables). Therefore, it is intuitive to modify the latent representation. By increasing the distance between the latent representations of the copyrighted and attack images, the embedded watermark is likely to be disrupted or removed.
Formally, we construct our intuitive objective function as:
\begin{equation}\label{eq:init_obj}
    \min_{\mathbf{x}^*_T} d_{\text{visual}}(\mathbf{x}^w_0, \mathbf{x}^*_0) - \gamma d_{\text{latent}}(\mathbf{x}_T, \mathbf{x}^*_T)
\end{equation}
where \(\gamma\) is a hyperparameter, \(d_{\text{visual}}\) and \(d_{\text{latent}}\) are distance metrics designed to maintain image similarity while encouraging differences in the latent space.

However, through preliminary experiments (see Appendix for more details), we found that this simple objective is difficult to achieve based on the following challenges:
\begin{itemize}
    \item \textbf{High Memory Consumption:} Optimizing Eq.~\ref{eq:init_obj} requires computing \(\frac{\partial \bv{x}_{t-1}^*}{\partial \bv{x}_{t}^*}\) at every timestep, consuming roughly 10 GB of GPU memory per step. Even a DPM solver~\cite{lu2022dpmSolver} with only 10 steps would demand over 100 GB, which far exceeds the memory limits of most GPUs.
    \item \textbf{Over-smoothing Images:} To reduce memory usage, some researchers estimate gradients by skipping several sampling steps~\cite{wang2024dpo}. However, this shortcut tends to over-smooth images. 
    \item \textbf{Noisy Outputs:} High perturbation in the latent space often results in noisy images rather than meaningful semantic alterations.
\end{itemize}





\section{Method}
\begin{figure}[t]
    \centering
    \includegraphics[width=0.5\textwidth]{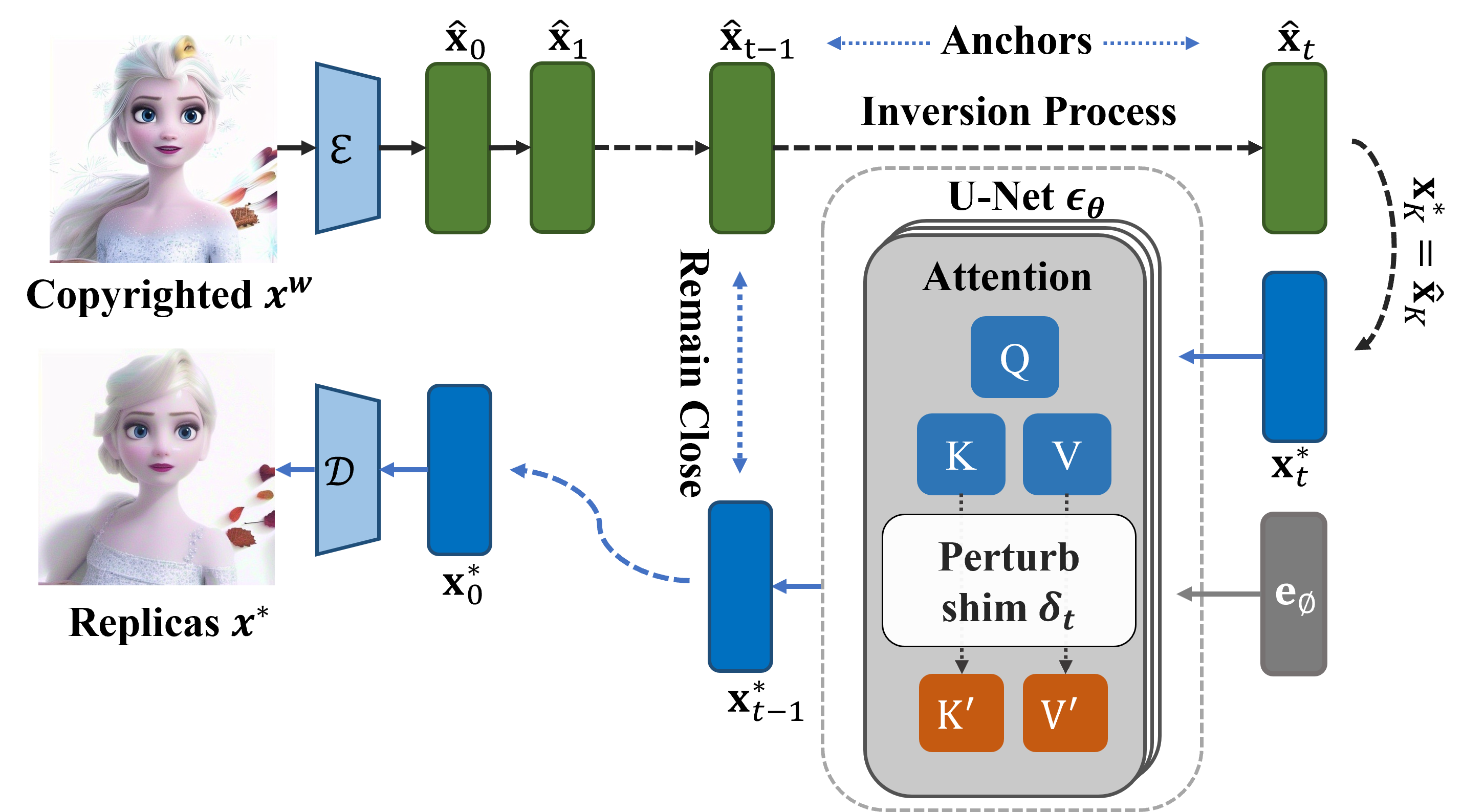}
    \caption{Attack pipeline overview for Neural plagiarism.}
    \label{fig:algorithm_overview}
\end{figure}

To address above challenges, we propose an optimization framework using \emph{anchors} and \emph{shims}, as shown in Figure~\ref{fig:algorithm_overview}.

\subsection{Optimization with Anchors and Shims}
We first obtain {\bf anchors} as a guideline for subsequent optimization. Given a target image $x$ (\ie, a copyrighted image $x^w$), we encode it into latent space via VAE encoder~\cite{Rombach_2022_CVPR}, obtaining the latent representation $\hat{\mathbf{x}}_0$. By inverting the generation process~\cite{hertz2022prompt} of a deterministic solver (\eg, DPM~\cite{lu2022dpmSolver}), we obtain a sequence of anchors as,
\begin{equation}
    \{\hat{\mathbf{x}}\}^T_{i=1} \coloneq \{\hat{\mathbf{x}}_1,..., \hat{\mathbf{x}}_T \}
\end{equation}

The above inverse latents can be used to generate semantically similar images. Here, $\hat{\mathbf{x}}_T$ is often used for latent watermark retrieval~\cite{tree_ring, yang2024gaussian}. Our intuitive objective is to find a perturbed $\mathbf{x}^*_T$ that is far away from $\hat{\mathbf{x}}_T$ (see eq~\ref{eq:init_obj}). However, since we only focus on replicating copyrighted image $x$, we can relax this goal to allow perturbation at arbitrary timestep $t$, as long as the resulting $\mathbf{x}^*_t$ yields satisfiable replication. Formally, we define this perturbation process as $\Delta(\cdot)$ along with a margin variable $\delta_t$,
\begin{equation}
    \mathbf{x}^*_{t} = \Delta_{\delta_t \in \mathcal{S}}(\mathbf{x}_t, \delta_t)
\end{equation} 
Notably, $\Delta$ is defined in a general form, indicating the perturbation could occur in any space $\mathcal{S}$ - not necessarily restricted to the latent space.
Instead to learn $\mathbf{x}^*_{t}$ directly, we find $\delta_t$, referred as the {\bf shim} during optimization. The norm of $\delta_t$ is referred as {\it the least} distance between the input and its perturbed result in the perturbation space $\mathcal{S}$.

The term ``anchors'' and ``shim'' originate from the alignment process in door installation, where a handyman inserts wooden shims at various locations to adjust spacing, ensuring uniform gaps to the anchors. Inspired by this concept, we perturb latent by adjusting $\delta_t$ with a loss term, 
\begin{equation}\label{eq:loss_norm}
    \mathcal{L}_{\text{norm}}(t) = \max(0,  \hat{\varepsilon}_t  - \| \delta_t \|)
\end{equation}
Here, the $\hat{\varepsilon}_t$ is the hyper-parameter to ensure a sufficient large distance. 

By adding shims to the latent variables, the resulting representations are likely to diverge from the anchors, thereby disrupting embedded watermarks and removing copyrighted information.

\subsection{Semantic Search via Attention Perturbation}
We aim to extract the semantically equivalent information from diffusion models to form a semantic search space when forging replicas. In stable diffusion, a conditional image generator is implemented by the cross-attention mechanism~\cite{vaswani2017attention} to align semantic prompts with images. By employing the UNet backbone and domain specific encoder $\tau_\theta$, the cross attention is implemented as $\text{Attention}(Q, K, V) = \text{softmax}\left(\frac{QK^T}{\sqrt{d_k}}\right) V$~\cite{Rombach_2022_CVPR}, where
\[
Q = W_Q^{(i)} \,\varphi_i(z_t), 
\quad K = W_K^{(i)} \,\tau_\theta(y), 
\quad V = W_V^{(i)} \,\tau_\theta(y).    
\]
Here, $W_Q^{(i)}$, $W_K^{(i)}$, and $W_V^{(i)}$ are learnable projection matrices; $\varphi_i(z_t)$ denotes a flattened intermediate representation of the U-Net implementing $\epsilon_\theta$; $\tau_\theta$ is the CLIP encoder~\cite{radford2021learning} to map a prompt $y$ into high-dimensional, dense text embeddings.

Our goal is to discover an alternative set of ($Q',K',V'$) that yield similar outputs to the original ones. At each diffusion timestep $t$, we inject ``shim'' $\delta_t$ on the text embedding only. 
To preserve semantic consistency, we maximize cosine similarity,
\begin{equation}\label{eq:loss_semantic}
    \mathcal{L}_{\text{semantic}}(t) = -\frac{ \mathbf{e}_{\emptyset} \cdot (\mathbf{e}_{\emptyset} + \delta_t)}
    {\|\mathbf{e}_{\emptyset}\| \|\mathbf{e}_{\emptyset} + \delta_t\|}
\end{equation}
Here, \(\mathbf{e}_{\emptyset}\) denotes the text embedding of an empty string, as encoded by the CLIP text encoder. This is a standard setting for unconditional generation and inversion~\cite{nichol2021glide}.

Meanwhile, to ensure the perturbed latents can produce similar outputs, we align the predicted next latent to remain close to its corresponding anchor, 
\begin{equation}\label{eq:loss_latent}
 \mathcal{L}_{\text{align}}(t) = d\left(\mathbf{x}_{t-1},  \hat{\mathbf{x}}_{t-1} \right)
\end{equation}
Specifically, we perturb with $\mathbf{x}^*_t=\Delta(\mathbf{x}_t, \delta_t)$ and remove the noise predicted by UNet: $\mathbf{x}_{t-1} \gets \mathbf{x}^*_t - \zeta_t \cdot \epsilon_{\theta}(\mathbf{x}^*_t, t)$ with coefficient $\zeta_t$ specified by different solvers.

\subsection{The Iterative Searching Process}
Equations~\ref{eq:loss_norm}, ~\ref{eq:loss_semantic}, ~\ref{eq:loss_latent} together define our new objective: {\it insert a large enough shim to induce semantic changes while keeping outputs remain close to anchors}. At a timestep $t$, $\delta_t$ is being searched jointly by,
\begin{equation}
    \min_{\delta_t}  \mathcal{L}_{\text{norm}}(t) + \gamma_1 \mathcal{L}_{\text{semantic}}(t) + \gamma_2 \mathcal{L}_{\text{align}}(t) 
\end{equation}
with $\gamma_1$ and $\gamma_2$ be the hyper-parameters.

Keen readers may have realized the major reason to introduce anchors and shims is to alleviate the aforementioned memory consumption issue. As we preserve the trajectory as anchors, we decouple the chain of latents, thereby we are free to adjust the shims at arbitrary timesteps — one by one.

Besides, the perturbation process can be done only on selected timesteps, improving the efficiency significantly. Notably, we add shims at specific timesteps, indicating that the semantic search space evolves over time. In particular, as $t$ decreases from $T$ to $1$, the shim gradually transitions from encoding large semantic changes to capturing finer semantic nuances. The attack can select a larger timestep to induce significant semantic alterations in copyrighted images or a smaller timestep to produce indistinguishable replicas that can evade invisible watermarks. Algorithm~\ref{alg:attack_pipeline} illustrates our overall attack pipeline proposed in this work.

\begin{algorithm}
\caption{The Attack Pipeline for Neural Plagiarism}\label{alg:attack_pipeline}
\begin{algorithmic}[1]
    \State \textbf{Input:} $x^w$
    \State \textbf{Require:} selected timesteps $\mathcal{T}_\text{select}$, perturbation function $\Delta(\cdot)$, learning rate $\eta$, coefficient $\zeta_t$\footnotemark[1], search depth $K$ with total timesteps $T$, VAE encoder $\mathcal{E}$ and decoder $\mathcal{D}$.
    
    \State $\hat{\mathbf{x}}_0 \gets \mathcal{E}(x^w)$
    \For{$t=1$ to $T$}
        \State $\hat{\mathbf{x}}_{t} \gets \text{DPMSolver}(\hat{\mathbf{x}}_{t-1})$ 
    \EndFor

    \State Initialize $\mathbf{x}^*_K = \hat{\mathbf{x}}_K$ or $\sqrt{\overline{\alpha}_K}\,\mathbf{x}_0 + \sqrt{1-\overline{\alpha}_K}\,\mathbf{\epsilon}$
    \For{$t=K$ to $1$}
        \If{$t \in \mathcal{T}_\text{select}$} \Comment{Enable gradient propagation}
            \State Initialize $\delta_t \gets \mathbf{0}$
            \While{$\delta_t$ not converged} 
                \State $\mathbf{x}^*_{t-1} \gets \mathbf{x}^*_t - \zeta_t \cdot \epsilon_{\theta}\left(\mathbf{x}^*_t, t, \mathbf{e}_{\emptyset} + \delta_t \right)$
                \State $\delta_t \gets \delta_t - \eta \nabla \mathcal{L}(t, \delta_t, \mathbf{x}^*_{t-1}, \hat{\mathbf{x}}_{t-1})$
            \EndWhile
        \Else \Comment{Disable gradient propagation}
            \State $\mathbf{x}^*_{t-1} \gets \mathbf{x}^*_t - \zeta_t \cdot \epsilon_{\theta}(\mathbf{x}^*_t, t, \mathbf{e}_{\emptyset})$
        \EndIf
    \EndFor

    \State \textbf{Output:} $x^* \gets \mathcal{D}(\mathbf{x}^*_0)$
\end{algorithmic}
\end{algorithm}
\footnotetext[1]{The coefficient $\zeta_t$ varies and is determined by different solvers. For example,\(\zeta_t = \Bigl(\sqrt{1-\alpha_t}-\sqrt{1-\alpha_{t-1}}\Bigr)\frac{\sqrt{\alpha_{t-1}}}{\sqrt{1-\alpha_{t-1}}}\) in DDIM~\cite{song2020denoising}.}

\section{Experiments}
\subsection{Setups}
\noindent\textbf{Datasets.} We select MS-COCO to evaluate our attack performance as the several watermarking methods~\cite{zhang2019robust, fernandez2023stable} utilize MS-COCO for training or fine-tuning. We randomly sample images from the MS-COCO dataset~\cite{lin2014microsoft} and crop them to $512\times512$ resolution. Additionally, we generate $512\times512$ images using corresponding prompts with Stable Diffusion (e.g., ``stable-diffusion-2-1-base"~\cite{Rombach_2022_CVPR}).

\begin{table*}[t]
\centering
\resizebox{.97\textwidth}{!}{%
\begin{tabular}{ccccccc|cccccc}
\toprule
\multirow{1}{*}{} & \multicolumn{6}{c}{\bf Late start, noisy latent} & \multicolumn{6}{c}{\bf Early start, inverse latent} \\
Method & BA $\downarrow$ & ACC $\downarrow$ & T{\@}1\%F $\downarrow$ & PSNR $\uparrow$ & SSIM $\uparrow$ & FID $\downarrow$ &
         BA $\downarrow$ & ACC $\downarrow$ & T{\@}1\%F $\downarrow$ & PSNR $\uparrow$ & SSIM $\uparrow$ & FID $\uparrow$ \\
\midrule
\multicolumn{13}{l}{\textbf{DwtDctSvd (32 bits)}} \\
Watermarked   & 1.00 & 1.00 & 1.00 & 38.33 & 0.99 & 3.54 & 1.00 & 1.00 & 1.00 & 39.48 & 0.99 & 3.02 \\
Regen         & 0.64 & 0.15 & 0.00 & 26.21 & 0.75 & 36.48 & 0.63 & 0.14 & 0.00 & 25.51 & 0.80 & 21.36 \\
Rinse         & 0.54 & 0.04 & 0.00 & 23.68 & 0.68 & 87.33 & 0.59 & 0.12 & 0.00 & 23.34 & 0.73 & 37.98 \\
\textbf{Ours} & 0.52 & 0.01 & 0.00 & 25.27 & 0.73 & 41.69 & 0.52 & 0.03 & 0.00 & 21.16 & 0.70 & 71.14 \\
\midrule
\multicolumn{13}{l}{\textbf{RivaGAN (32 bits)}} \\
Watermarked   & 1.00 & 1.00 & 1.00 & 40.57 & 0.98 & 3.81 & 1.00 & 1.00 & 1.00 & 40.60 & 0.98 & 3.62 \\
Regen         & 0.60 & 0.05 & 0.00 & 26.21 & 0.75 & 34.85 & 0.59 & 0.01 & 0.00 & 25.40 & 0.79 & 21.27 \\
Rinse         & 0.52 & 0.01 & 0.00 & 23.74 & 0.68 & 82.72 & 0.53 & 0.01 & 0.00 & 23.42 & 0.73 & 36.98 \\
\textbf{Ours} & 0.56 & 0.02 & 0.00 & 25.29 & 0.73 & 40.80 & 0.52 & 0.00 & 0.00 & 21.15 & 0.70 & 71.14 \\
\bottomrule
\end{tabular}%
}
\caption{Watermark removal for invisible post-hoc watermarking on COCO images. A lower FID is preferred when removing invisible watermarks and a higher FID is preferred to encourage semantic changes for altering copyrighted images.}
\label{tbl:invisible_watermark_removal}
\end{table*}

\noindent\textbf{Search Configuration.} By default, we adopt the Adam optimizer~\cite{kingma2015adam} with a learning rate of 0.01. To stabilize the optimization process, we utilize weight decay (i.e., L2 regularization, set to $10^{-3}$) to balance the shims and apply gradient norm clipping (with a maximum norm of 1.0) to prevent exploding gradients. 
We use DPM solver~\cite{lu2022dpmSolver} with a scheduler of 50 timesteps. For hyper-parameters, we set $\gamma_1=10^5$, $\gamma_2=0.1$ and $\hat{\varepsilon}_t=10$ for every timestep.


\subsection{Forgery Attack on Copyrighted Images}
Forgery attacks aim to generate replicated images that bypass copyright protections. We investigate two scenarios: attacks on copyrighted images and invisible watermark removal. We compare our approach against two baseline regeneration attacks. Regen~\cite{zhao2023invisible} adds noise at a specific timestep and then immediately denoises, while Rinse~\cite{an2024benchmarking} repeats this process multiple times (e.g., twice). Notably, all experiments employ the same model architecture and weights, without additional components such as extra image encoders or ControlNet adapters, differing only in the generation pipeline.

{
\setlength{\tabcolsep}{0.35pt}
\begin{table*}[ht]
\centering
\begin{tabular}{ccccccc}
initial & iter 0 & iter 1 & iter 2 & iter 4 & iter 8 & attack\\
\midrule
\multicolumn{7}{c}{\small\textbf{Early start, noisy latent}} \\
\includegraphics[width=0.14\linewidth]{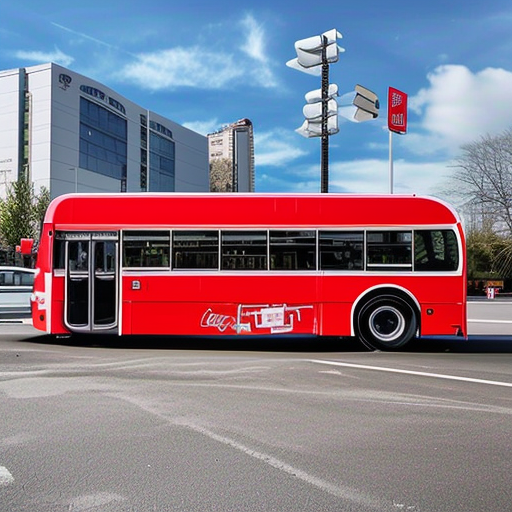} & 
\includegraphics[width=0.14\linewidth]{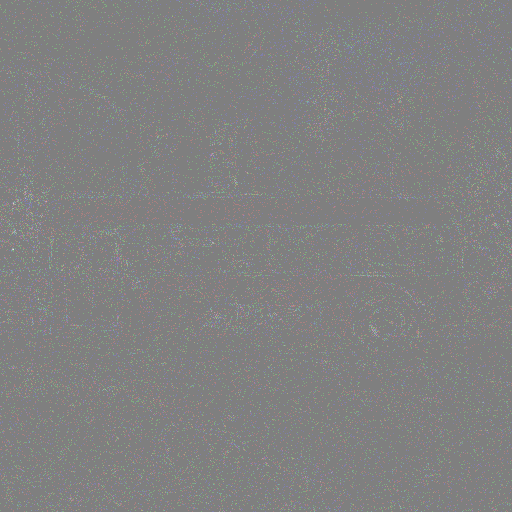} & 
\includegraphics[width=0.14\linewidth]{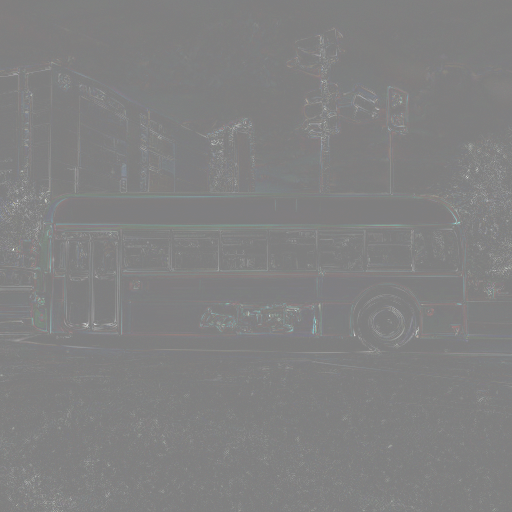} & 
\includegraphics[width=0.14\linewidth]{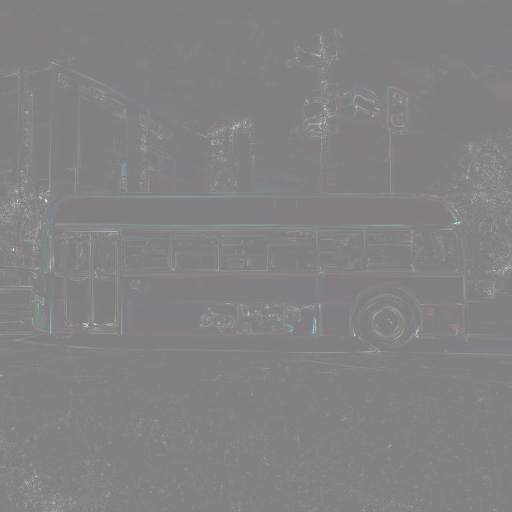} & 
\includegraphics[width=0.14\linewidth]{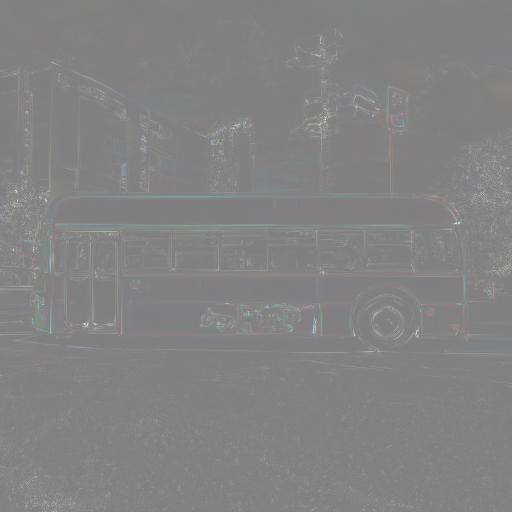} & 
\includegraphics[width=0.14\linewidth]{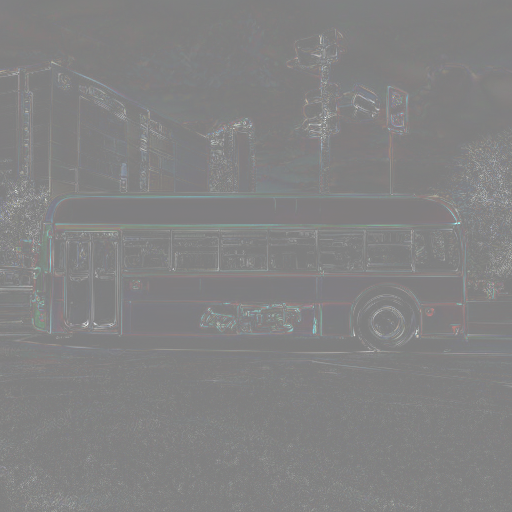} & 
\includegraphics[width=0.14\linewidth]{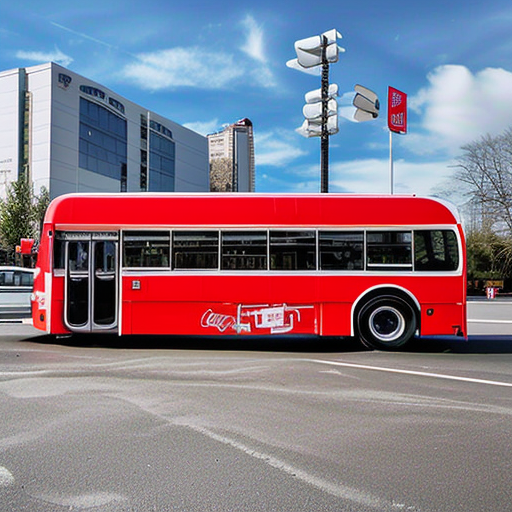} \\

\multicolumn{7}{c}{\small\textbf{Late start, inverse latent}} \\
\includegraphics[width=0.14\linewidth]{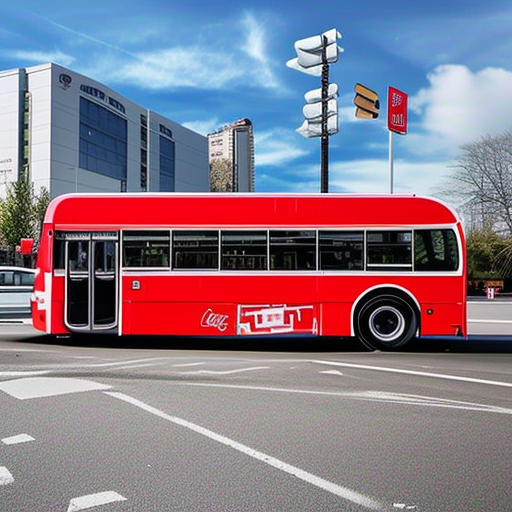} & 
\includegraphics[width=0.14\linewidth]{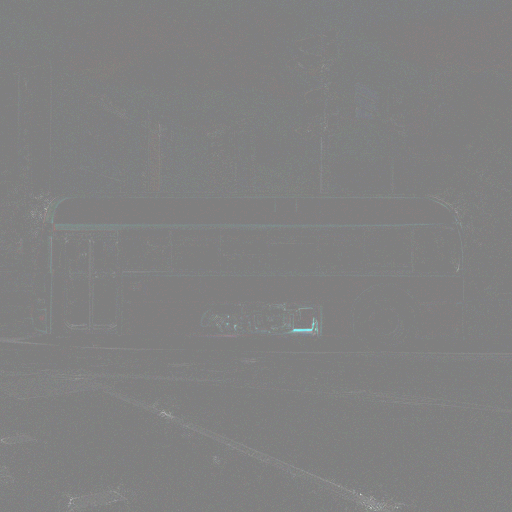} & 
\includegraphics[width=0.14\linewidth]{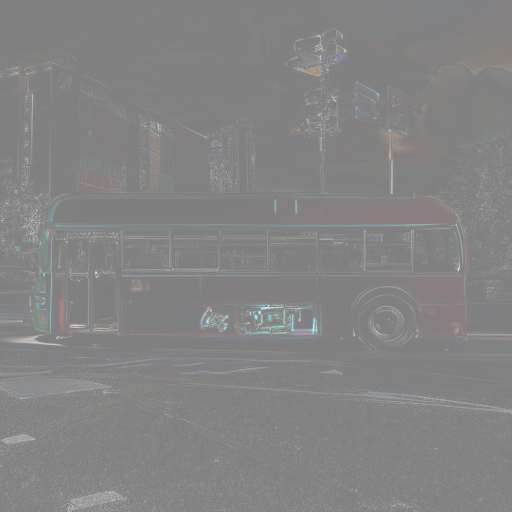} & 
\includegraphics[width=0.14\linewidth]{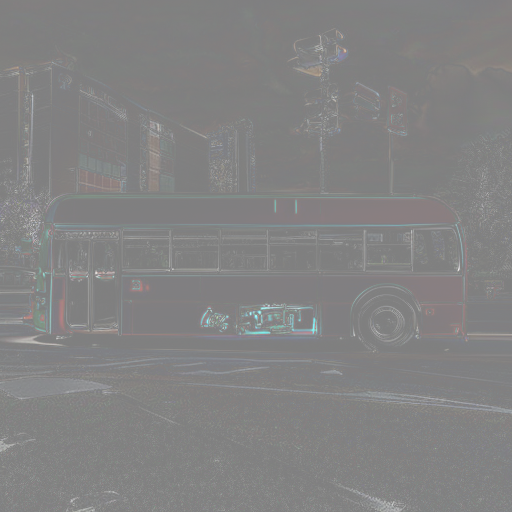} & 
\includegraphics[width=0.14\linewidth]{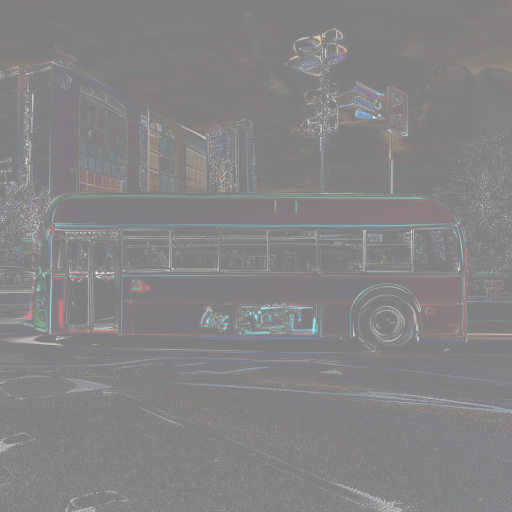} & 
\includegraphics[width=0.14\linewidth]{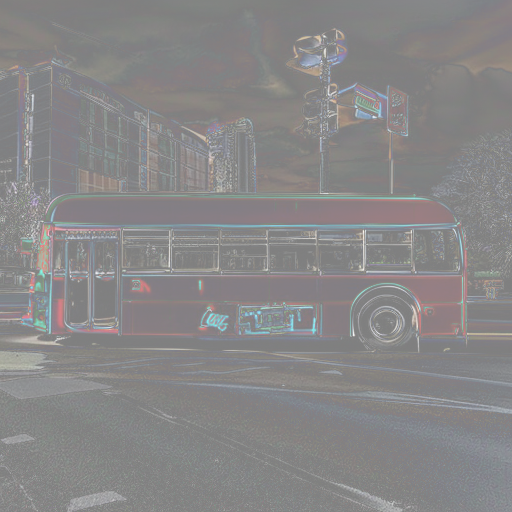} & 
\includegraphics[width=0.14\linewidth]{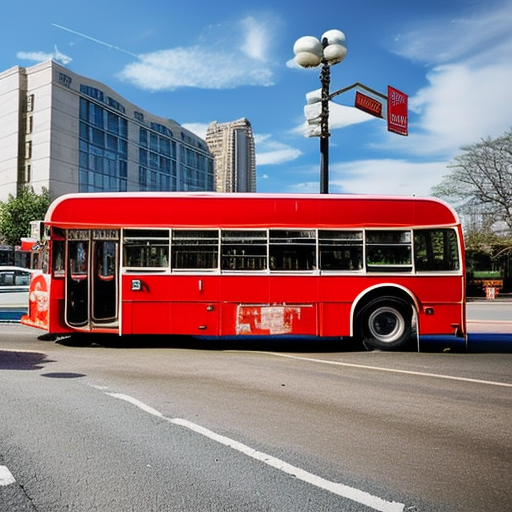} \\

\end{tabular}
\caption{The middle columns highlight the pixel differences between the initial image (leftmost) and the attack image (rightmost) during optimization. These illustrate how the latent gradually diverges from the anchor latent in a seamless and semantically meaningful manner.}\label{tbl:diff_change}
\label{tab:image_comparison}
\end{table*}
}

\subsubsection{Invisible Watermark Removal}
We use the benchmark~\cite{zhao2023invisible} to evaluate our method on invisible watermark removal. Particularly, we employ performance metrics including: the changes in bit accuracy (BA), accuracy (ACC) defined as more than $\frac{2}{3}$ bits correctly identified, and the True Positive Rate when False Positive Rate below $1\%$ ($T@1\%F$), which is commonly recommended by the benchmark~\cite{zhao2023invisible, carlini2022membership}; For image generation quality and semantic similarity, we measure Peak Signal-to-Noise Ratio (PSNR)~\cite{gonzalez2002digital}, Structural Similarity Index (SSIM)~\cite{ndajah2010ssim} and Fr\'{e}chet Inception Distance (FID)~\cite{heusel2017gans}. 

\noindent\textbf{Post-hoc Watermarking Methods:} we explore two common post-hoc watermarking methods,   DctDwtSvd~\cite{hamidi2021dct} and RivaGAN~\cite{zhang2019robust}, which are applied on both real images and generated images using MS-COCO.

As in Algorithm~\ref{alg:attack_pipeline}, our attack starts at timestep \(K\) and backward step by step until reaching timestep \(1\). An early start indicates that we begin at a large timestep, while a late start corresponds to beginning at a smaller timestep.
\begin{itemize}
    \item \textbf{Late start with noisy latent:} small perturbations are effective for removing invisible watermarks while preserving the original image quality. To achieve this, we begin our attack at later timestep ($K=140$) and insert shims at selected timesteps $100$ and $60$. For this setting, we use a noisy start defined as $\mathbf{x}_K^*=\sqrt{\overline{\alpha}_K}\,\mathbf{x}_0 + \sqrt{1-\overline{\alpha}_K}\,\mathbf{\epsilon}$ since we observed that starting with inverse latent leads to limited semantic changes and is ineffective for invisible watermark removal. Consequently, although our approach shares the same starting latent as Regen~\cite{zhao2023invisible}, our approach works much better due to the perturbation shims  of our method. As shown in Table~\ref{tbl:invisible_watermark_removal} (left column), our method achieves significantly improved watermark removal performance, close to $50\%$ bit accuracy, while maintaining moderate PSNR, SSIM, and FID.
    \item \textbf{Early start with inverse latent:} large perturbations cause dramatic alterations to image features, which may remove the watermark in the image. We explore this settings with beginning time step $K=1000$ and insert shims at selected timesteps $600$ and $200$ to induce a large perturbation on attack images. As shown in Table~\ref{tbl:invisible_watermark_removal} (right column), the detection bit accuracy also drops to nearly $50\%$, indicating successful watermark removal. However, this approach significantly alters the feature distribution, resulting in much higher FID scores compared to other settings. Notably, a high FID here does not necessarily imply that the attacked images are of low quality or noisy, as our method encourages meaningful semantic changes instead to random noise.
\end{itemize}

\textbf{Why these settings?} In Table~\ref{tbl:diff_change}, we visualize the pixel differences between the target images (watermarked images) and the attack images to illustrate the semantic changes induced by the shims. When we use the inverse latent as startup latent at a large timestep (early start), it induces significant semantic changes, for example, altering color of a bus, shape of traffic lights or the design of windows. In contrast, starting with the inverse latent at a small timestep (late start) produces minor and localized semantic changes, as major semantic changes have already been handled in previous timesteps. In this case, we use a noise latent to force the search for all detailed semantic changes, thereby improving the performance for invisible watermark removal (See Appendix for visualizations of all other cases.). 

\noindent\textbf{Latent Watermarking Methods:} we also investigate two methods that apply watermarking in the latent space: Stable Signatures~\cite{fernandez2023stable} and Tree-Ring~\cite{tree_ring}. Table~\ref{tbl:invisible_watermark_removal2} shows the results. It is clear that because the shims alter the latent variables, they generate alternative image latents that can bypass Stable Signatures, a watermarking method that fine-tunes the VAE decoder. However, our method is unable to remove the Tree-Ring watermark, as it injects a pattern on latent variable in Fourier space using masking; even large shifts in the latent space do not significantly alter the local Fourier characteristics.

\begin{table}[h]
\centering
\resizebox{0.45\textwidth}{!}{%
\begin{tabular}{ccccc}
\toprule
Method & T@1\%F $\downarrow$ & PSNR $\uparrow$ & SSIM $\uparrow$ & FID $\downarrow$ \\
\midrule
\multicolumn{5}{l}{\textbf{Stable Signature}} \\
Watermarked & 1.00  & 11.79 & 0.38 & 153.14  \\ 
Regen & 0.00 & 24.94 & 0.78 & 24.91  \\
Rinse & 0.00 & 22.96 & 0.72 & 46.85 \\
\textbf{Ours} & 0.00 & 24.08 & 0.80 & 43.83 \\ 
\midrule
\multicolumn{5}{l}{\textbf{Tree-ring}} \\
Watermarked & 1.00  & 12.99 & 0.49 & 122.20  \\
Regen & 1.00  & 26.79 & 0.81 & 22.73  \\
Rinse & 1.00  & 23.68 & 0.75 & 39.96  \\
\textbf{Ours} & 0.98  & 25.84 & 0.86 & 33.50  \\
\bottomrule
\end{tabular}
}
\caption{Watermark removal for diffusion-based watermarking.}
\label{tbl:invisible_watermark_removal2}
\end{table}
However, this result does not mean that Tree-Ring is practical for copyright protection. Tree-Ring suffers from ambiguity attacks. In particular, the optimal threshold chosen by the $T@1\%F$ metric can create confusion among different watermark keys, causing images to be falsely classified as watermarked even when they were not watermarked with this key~\cite{ci2024ringid}.

\subsubsection{Copyrighted Images Replicas} 
We collect a set of online copyrighted images that include the common practices for copyright protection. The results are summarized in Table~\ref{tbl:copyright_image_replicates}. To evaluate the degree of semantic modification, we apply shims on various timesteps. 

It is clear that Regen~\cite{zhao2023invisible} and Rinse~\cite{an2024benchmarking} can partially remove copyright content. Despite of significant noise introduced, the resulting images remain similar enough to the originals, which would very likely be disputed as clear plagiarism. In contrast, our method can easily generate replicas while bypassing the copyrights restriction. For example, Elsa's iconic ``sophisticated braided updo'' hairstyle and `` icy elegance layer skirt'' are being replaced with short hair and suspender skirt, creating a blurring line for the Disney copyrighted anime. Additionally, we produce replicas for for Elon Musk, featured a thinner face with noticeable structural changes, which are uncommon in traditional AI deepfake. We also present Elon100, with 100 distinct replicas of Elon's portrait, none of which can be definitively identified as him by a GPT arbiter. (See appendix for more details)

\textbf{Neural Plagiarism in Real-world:} attackers can choose different timesteps or shims within the proposed attack pipeline, generating a large amount of replicas using random seeds. When targeting copyrighted images that feature visible trademarks, signatures, or IPs, they can cherry-pick the best replicas to bypass the copyright restrictions. Besides, the pipeline accepts both prompt inputs and negative prompts, the final attacks can be even more potent in real-world scenarios.


{
\setlength{\tabcolsep}{0.35pt}
\begin{table*}[t]
\centering
\begin{tabular}{ccccccc}
Target & Regen & Rinse & t=100 (ours) & t=300 (ours) & t=500 (ours) &t=700 (ours)\\
\midrule
\multicolumn{7}{c}{\small\textbf{Copyrighted IP: Frozen by Disney}} \\
\includegraphics[width=0.14\linewidth]{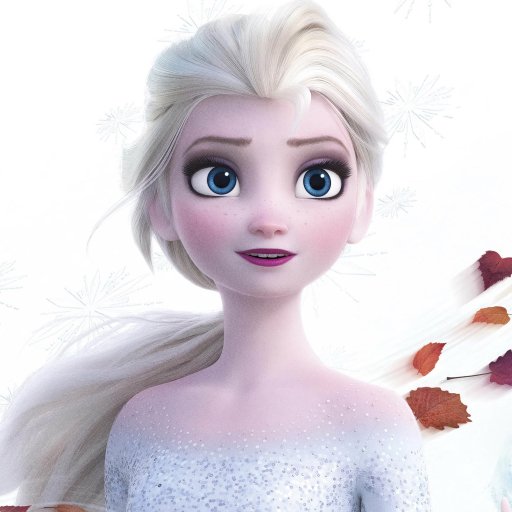} & 
\includegraphics[width=0.14\linewidth]{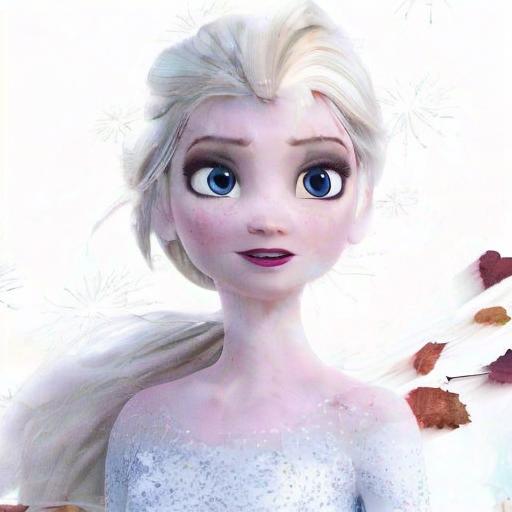} & 
\includegraphics[width=0.14\linewidth]{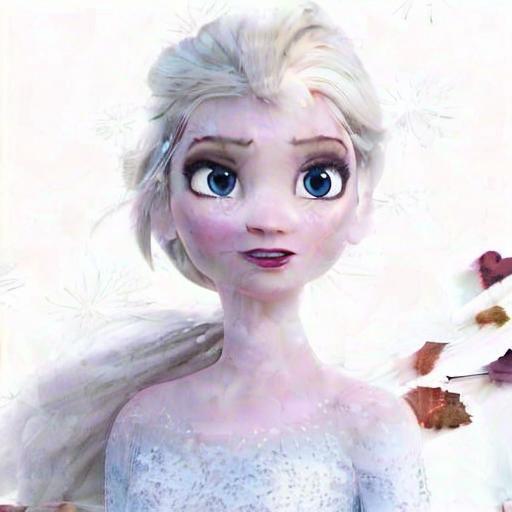} & 
\includegraphics[width=0.14\linewidth]{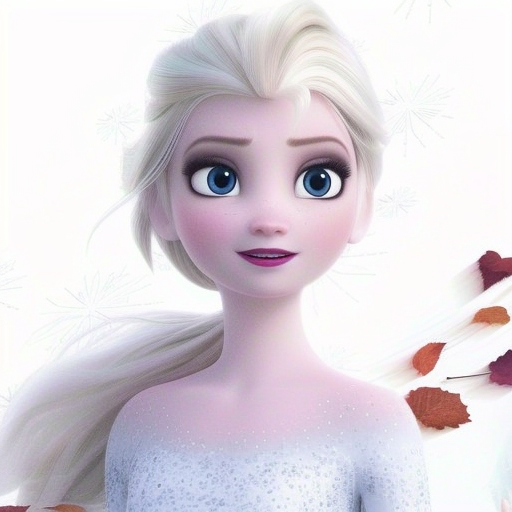} & 
\includegraphics[width=0.14\linewidth]{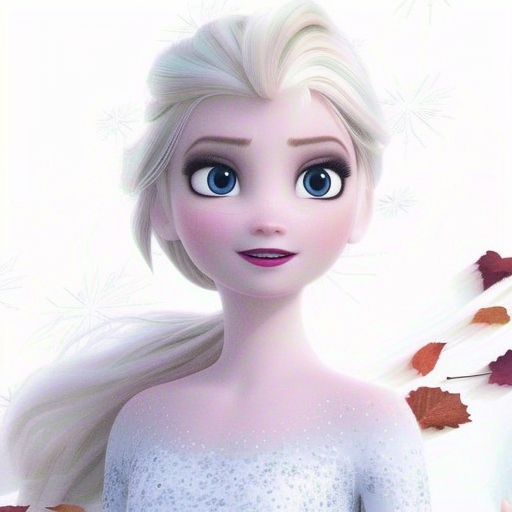} & 
\includegraphics[width=0.14\linewidth]{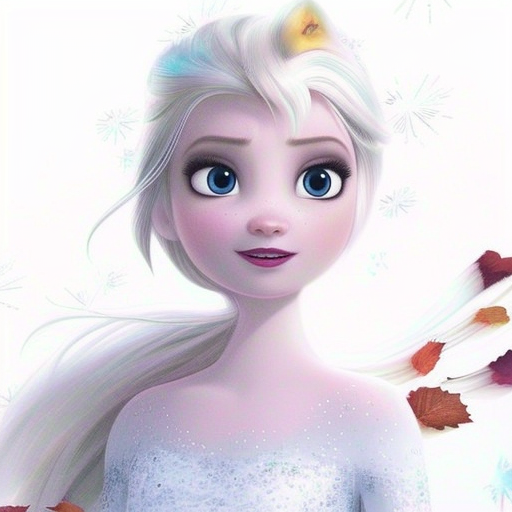} & 
\includegraphics[width=0.14\linewidth]{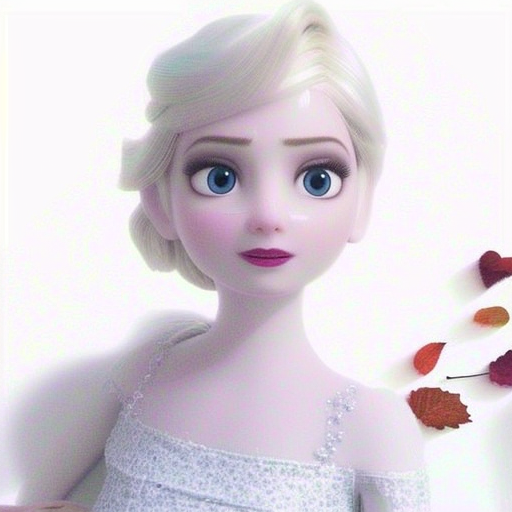} \\

\multicolumn{7}{c}{\small\textbf{Artwork: Claude Monet's impression painting}} \\
\includegraphics[width=0.14\linewidth]{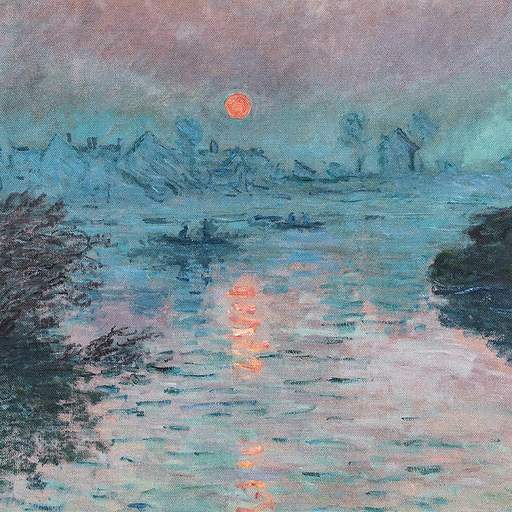} & 
\includegraphics[width=0.14\linewidth]{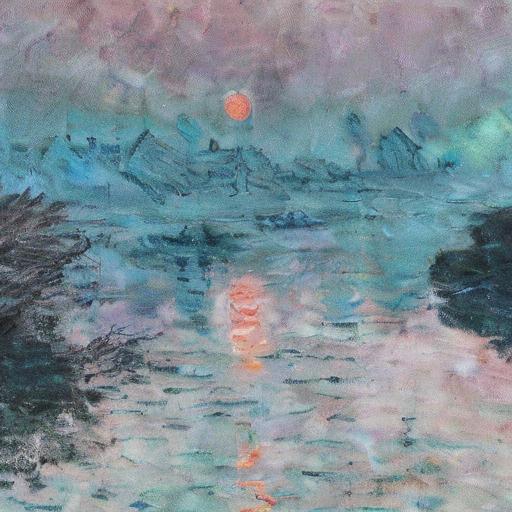} & 
\includegraphics[width=0.14\linewidth]{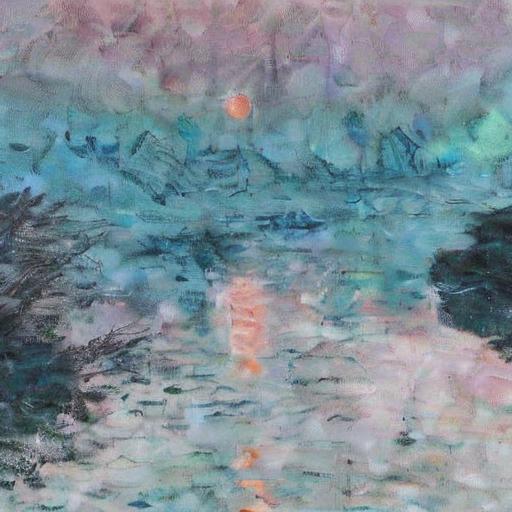} & 
\includegraphics[width=0.14\linewidth]{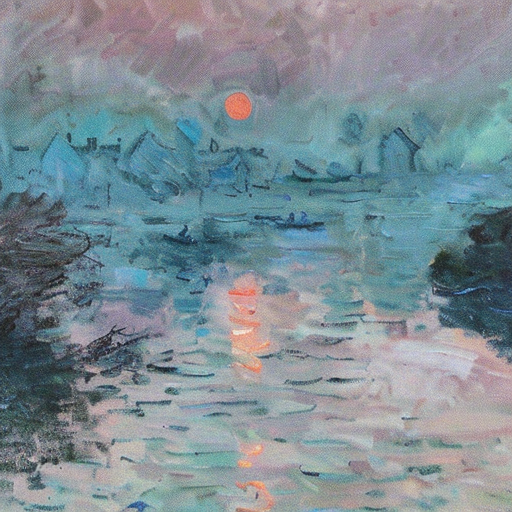} & 
\includegraphics[width=0.14\linewidth]{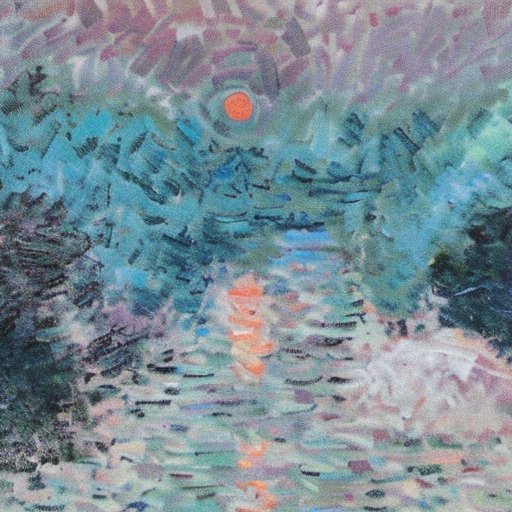} & 
\includegraphics[width=0.14\linewidth]{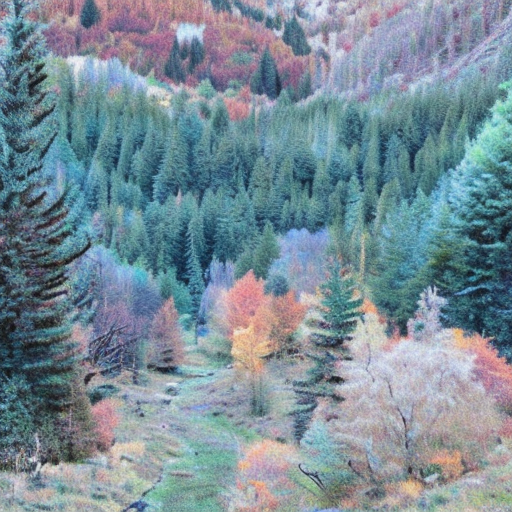} & 
\includegraphics[width=0.14\linewidth]{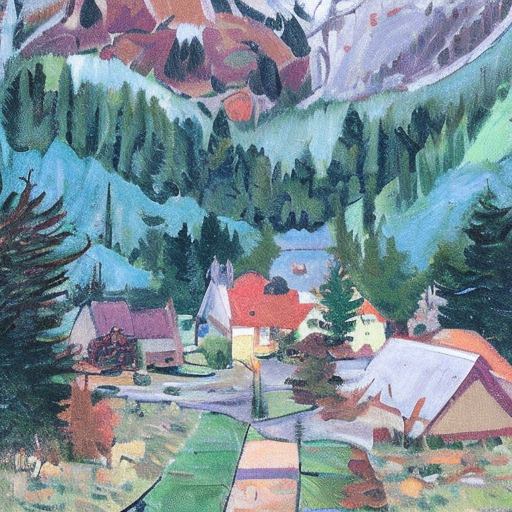} \\

\multicolumn{7}{c}{\small\textbf{Signature: Vincent van Gogh's Signature}} \\
\includegraphics[width=0.14\linewidth]{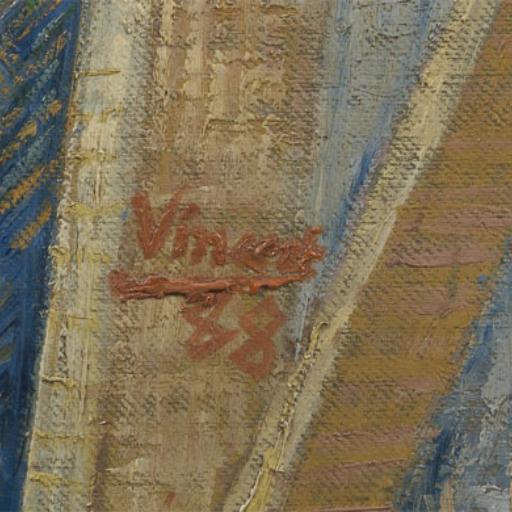} & 
\includegraphics[width=0.14\linewidth]{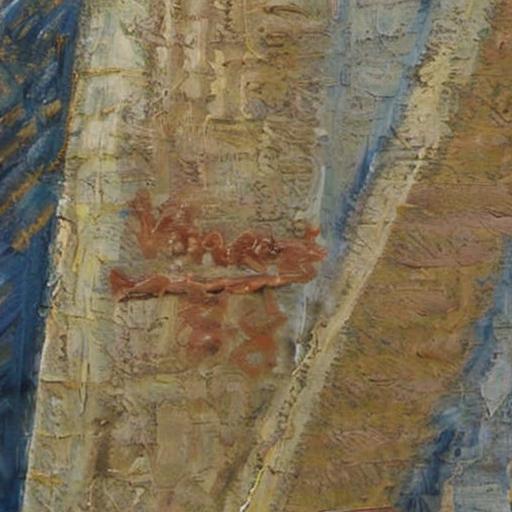} & 
\includegraphics[width=0.14\linewidth]{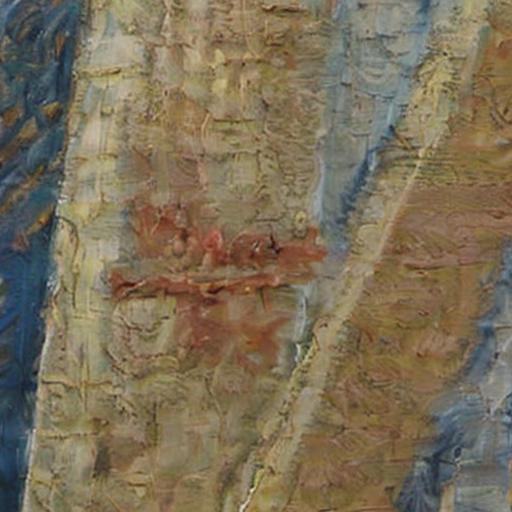} & 
\includegraphics[width=0.14\linewidth]{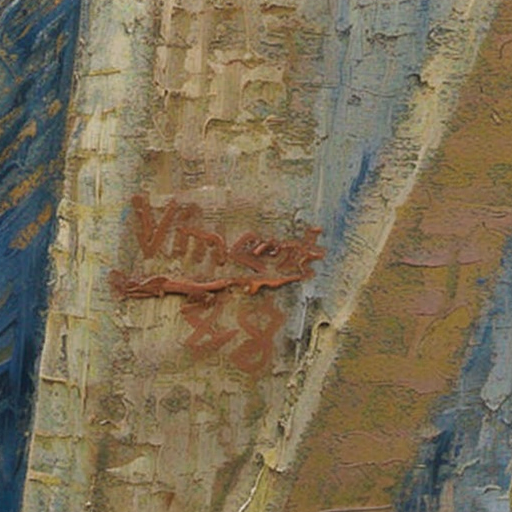} & 
\includegraphics[width=0.14\linewidth]{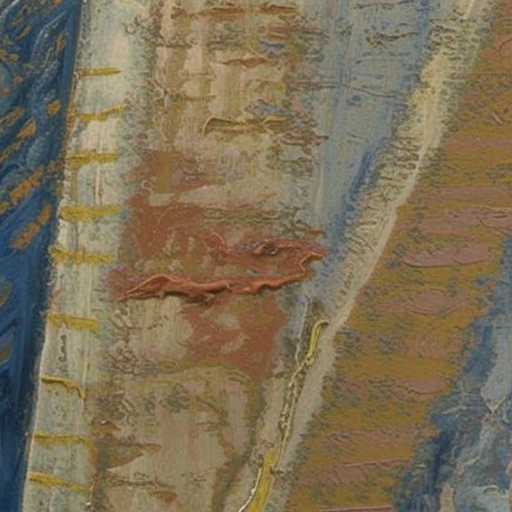} & 
\includegraphics[width=0.14\linewidth]{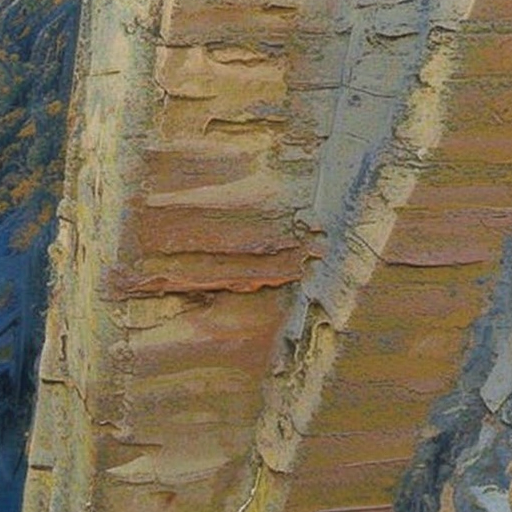} & 
\includegraphics[width=0.14\linewidth]{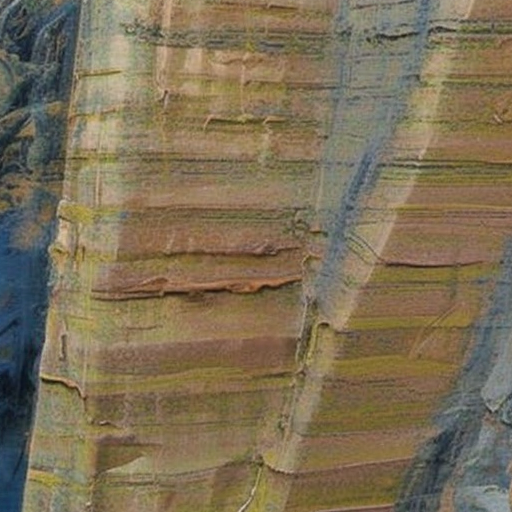} \\

\multicolumn{7}{c}{\small\textbf{Identity Portrait: Elon Musk}} \\
\includegraphics[width=0.14\linewidth]{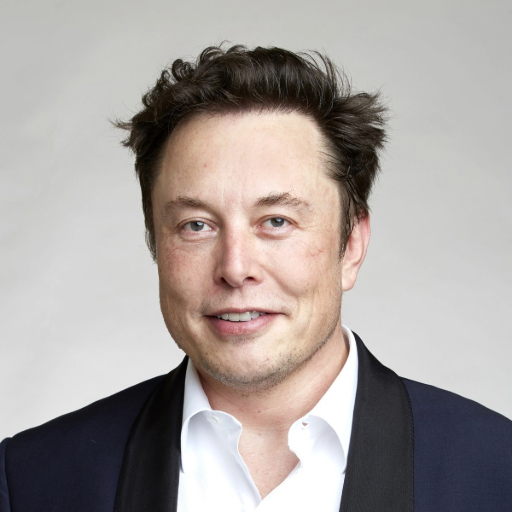} & 
\includegraphics[width=0.14\linewidth]{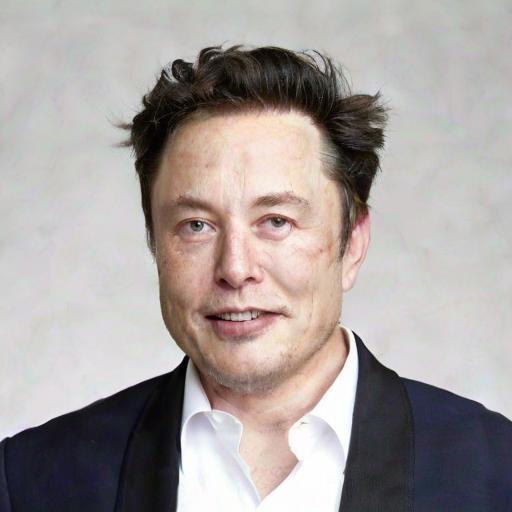} & 
\includegraphics[width=0.14\linewidth]{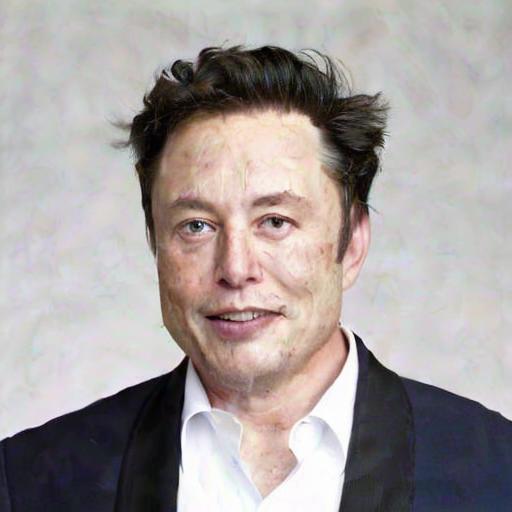} & 
\includegraphics[width=0.14\linewidth]{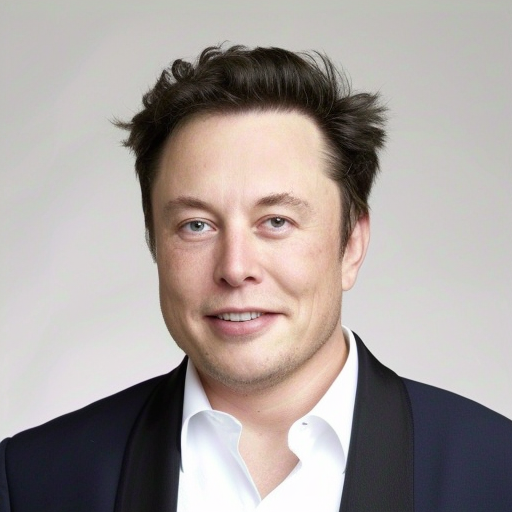} & 
\includegraphics[width=0.14\linewidth]{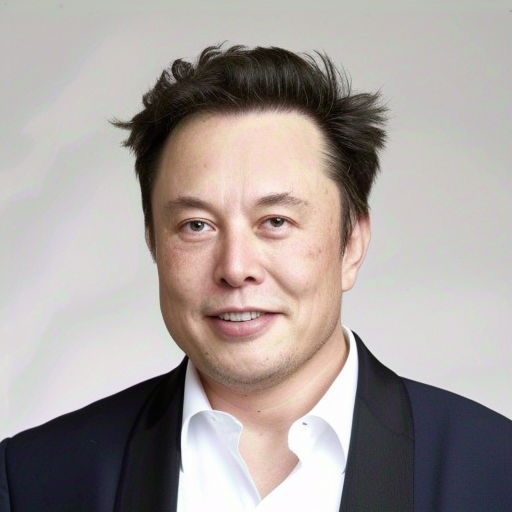} & 
\includegraphics[width=0.14\linewidth]{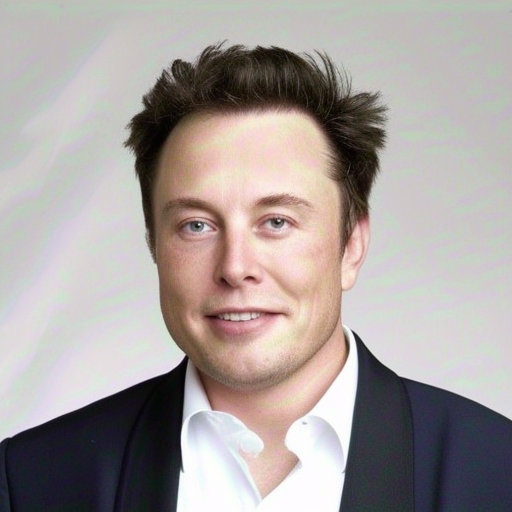} & 
\includegraphics[width=0.14\linewidth]{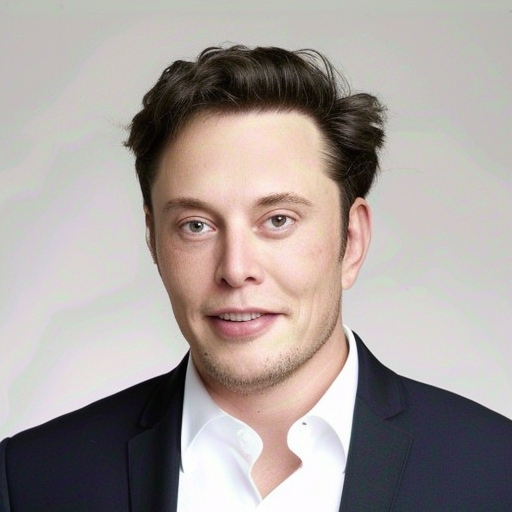} \\
\end{tabular}
\vspace{-5.5pt}
\caption{Plagiarize copyrighted images.}\label{tbl:copyright_image_replicates}
\label{tab:image_comparison}
\end{table*}
}
\vspace{-2.5pt}

\subsection{Ambiguity Attacks}
This is one of the most significant threat to copyright protection and has not received enough attention in the research community. 
We present two primary attack strategies:
\begin{itemize}
    \item \textbf{Watermark Replacement:} In the case of RivaGAN, we remove the original watermark via our attack pipeline and subsequently inserting a new watermark $w^*$. This watermark replacement effectively nullifies the original copyright protection and thereby introduce copyright ambiguity with new watermark. As shown in Table~\ref{tbl:ambiguity_attack}, the new watermark $w^*$ receives higher bit accuracy than the original one.
    \item \textbf{Co-existing Watermarks:} In the case of Tree-Ring, a latent watermark is injected into the inverse latent $\mathbf{x}^*_T$. Without precise knowledge of the channel containing the original watermark, the new watermark is embedded in an alternative channel, leading to the coexistence of both watermarks. From Table~\ref{tbl:ambiguity_attack}, both watermark receive high $T@1\%F$, creating ambiguity of data ownership.
\end{itemize}
\begin{table}[h]
\centering
\resizebox{0.47\textwidth}{!}{%
\begin{tabular}{ccccc}
\toprule
Method & BA ($w$) $\downarrow$ & BA ($w^*$) $\uparrow$ & T@1\%F ($w$) $\downarrow$ & T@1\%F ($w^*$) $\uparrow$ \\
\midrule
\multicolumn{5}{l}{\textbf{RivaGAN (32 bits)}} \\
Watermarked & 1.00 & 0.75 & 1.00 & 0.00 \\ 
Attacked & 0.75  & 1.00 & 0.00 & 1.00  \\
\midrule
\multicolumn{5}{l}{\textbf{Tree-ring}} \\
Watermarked & - & - & 1.00 & 0.00 \\
Attacked & - & - & 0.99 & 1.00  \\
\bottomrule
\end{tabular}}
\caption{Attack performance for ambiguity attacks. New watermark $w^*$ can be successfully recovered on the attacked images, introducing ambiguity for data copyright.}\label{tbl:ambiguity_attack}
\end{table}
\vspace{-2.5pt}

\section{Limitations and Future Work} 
Our method can remove most of watermark except for tree-ring; however, we can confuse the ownership detection successfully with alternative watermark key. 

Some approaches could be used to enhance our attacks, including advanced image editing techniques~\cite{hertz2022prompt}, integration with plugins such as ControlNet adapters~\cite{liu2024image}, image encoders~\cite{oquab2023dinov2}, and negative prompts~\cite{negativePrompts2022, negativeprompts2024}. We leave these as future work.

\section{Conclusion}
In this paper, we introduce a new threat, termed ``neural plagiarism'', in which diffusion models plagiarize copyrighted images while bypassing protections like watermarking. We propose a universal attack pipeline that enables both forgery and ambiguity attacks, resulting in effective data plagiarism. Withe extensive experiments, our approach is effective to evade copyright protections ranging from visible copyrighted images to invisible watermarking. 

\noindent\textbf{Acknowledgments:}  This work was supported in part by NSF grants 1952792 and 2321572, and Google. 

{\small
\bibliographystyle{ieeenat_fullname}
\bibliography{main}
}

\end{document}